%% file: main.tex
\documentclass[letterpaper,10pt,conference]{ieeeconf}

\IEEEoverridecommandlockouts
\usepackage{amsmath}
\usepackage{amssymb}
\usepackage{booktabs}
\usepackage{graphicx}
\usepackage{multirow}
\usepackage{array}
\usepackage{xcolor}
\usepackage{cite}
\usepackage{flushend}
\usepackage{hyperref}
\usepackage{url}
\title{\LARGE \bf
VICAR: Video Imitation with Contact-Aware Retargeting for Humanoid Interactive Skills
}

\author{Dvij Kalaria \ \ \ \ Bike Zhang \ \ \ \ S. Shankar Sastry\\ University of California, Berkeley}

\newcommand{\method}{VICAR}
\newcommand{\gmini}{GPT-4o mini}
\newcommand{\R}{\mathbb{R}}
\newcommand{\SO}{\mathrm{SO}}
\newcommand{\SE}{\mathrm{SE}}
\newcommand{\GL}{\mathrm{GL}}

\begin{document}
\maketitle
\thispagestyle{empty}
\pagestyle{empty}

\begin{abstract}
Third-person human videos offer a scalable alternative to task-specific reward engineering, teleoperation, and manually scripted demonstrations for humanoid loco-manipulation. Recent video-to-humanoid systems have shown that noisy human motion, object trajectories, and contacts can be extracted from unstructured videos and refined into deployable skills. However, a key gap remains for contact-rich imitation: the task is often defined by world-frame contacts whose poses, timing, and speed must survive embodiment mismatch and runtime scene changes. Towards this, we present \method{} (Video Imitation with Contact-Aware Retargeting), a pipeline for learning humanoid interactive skills from third-person human videos. The system first recovers human motion, reconstructs task-relevant obstacles from monocular video, and then retargets the motion to a humanoid while preserving global contact poses, motion speed, and collision clearance. Unlike retargeting methods that only remove penetration or preserve local interaction geometry, our formulation represents user-specified contact poses and collision clearance as high-priority Lagrangian penalty terms and optimizes the full motion to remain smooth through contact-rich transitions. The resulting motions are used to train a contact-conditioned motion generator whose inputs include movable contact points enabling runtime adaptation to new contact positions. We demonstrate the framework on ten diverse third-person video skills, including six table-tennis serves, three object-pickup tasks, and ladder climbing, and deploy the learned skills on hardware. These results suggest that contact-preserving retargeting is a practical step toward humanoids that learn useful interactive behavior by watching humans.

\textbf{Project website: } \href{https://ember-center-berkeley.github.io/VICAR}{ember-center-berkeley.github.io/VICAR}
\end{abstract}

\input{sections/01_introduction}
\input{sections/02_related_work}
\input{sections/03_method}
\input{sections/04_experiments}
\input{sections/05_limitations_conclusion}

% Do not add acknowledgments to the double-anonymous submission.
\bibliographystyle{IEEEtran}
\bibliography{references}

% ICRA 2027 counts appendices in the eight-page submission limit. The long
% appendix is therefore excluded from the submission build by default. Only
% enable these lines if the complete PDF, including references, remains at or
% below eight pages.
% \appendix
% \input{sections/appendix}

\end{document}

%% file: sections/01_introduction.tex
\section{Introduction}

Humans routinely learn interaction skills by watching other humans. A child does not need an adult to physically guide their body through every action, in the way teleoperation is often used to train robots; they can learn by observing that a cup should be reached on the table, that the table edge can support the body while crouching, or that a tossed ball must be met at a precise point in space. For humanoid robots, however, this mode of learning remains difficult: third-person videos provide abundant motion cues, but the robot must execute them with a different morphology, different joint limits, and different contact geometry. These tasks are organized around sparse contact events: a hand must arrive at an object, a foot must land on a rung, or a racket must meet a ball at the right pose and time. A trajectory can therefore resemble the source motion yet remain unusable if those events drift in the world frame.

% \begin{figure}
%     \centering
%     \includegraphics[width=\linewidth]{figures/top_rebuttal.pdf}
%     \caption{Temporally aligned human input and robot execution frames for representative skills: (a) table-tennis serving, (b) bimanual pick-and-place, (c) under-table pickup with table-edge support, and (d) ladder climbing in simulation. All tasks except ladder climbing are evaluated on hardware.}
%     \label{fig:top}
% \end{figure}

\begin{figure*}[t]
    \centering
    \includegraphics[width=0.97\textwidth]{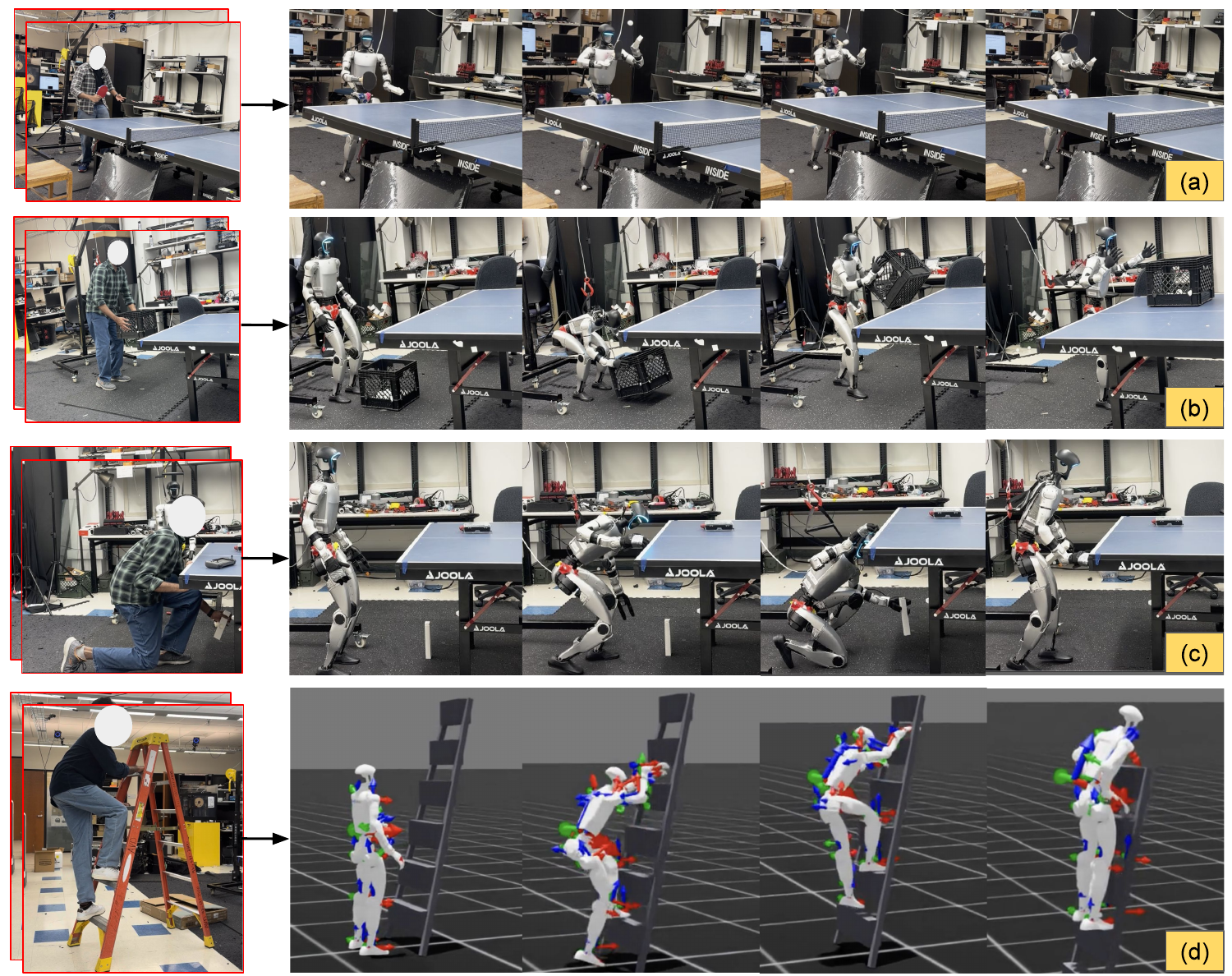}
    \caption{Temporally aligned human input and robot execution frames for representative skills: (a) table-tennis serving, (b) bimanual pick-and-place, (c) under-table pickup with table-edge support, and (d) ladder climbing in simulation. All tasks except ladder climbing are evaluated on hardware.}
    \label{fig:top}
\end{figure*}

\begin{figure}[t]
    \centering
    \includegraphics[width=0.55\linewidth]{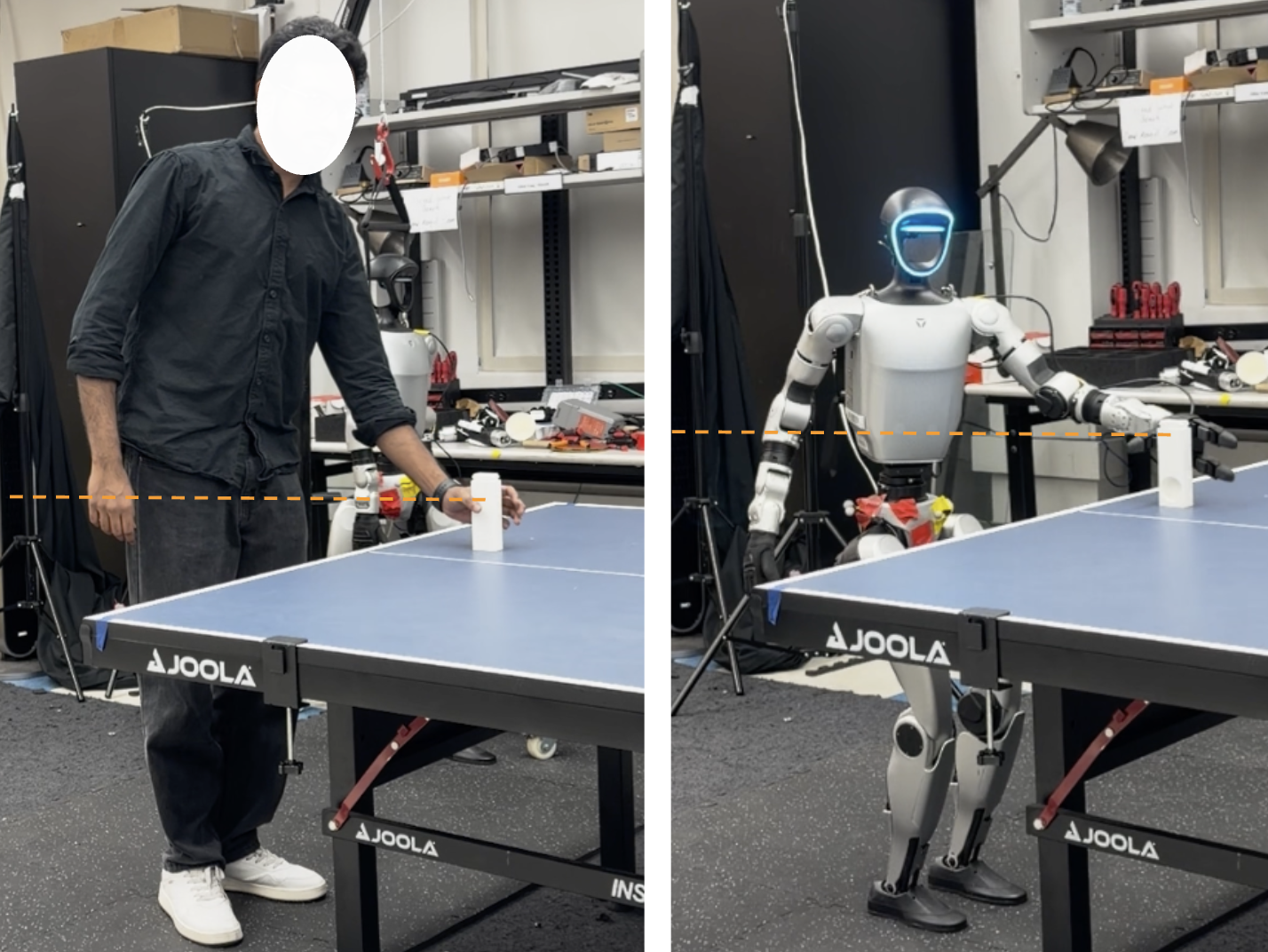}
    \caption{Relative table-height mismatch across embodiments: the same table is near waist height for the human but near chest height for the shorter G1.}
    \label{fig:embodiment_gap}
\end{figure}

The embodiment gap becomes especially severe for contact-rich tasks. Consider a standing adult picking up an object from a table. For the adult, the table may be near waist height; for a shorter humanoid such as Unitree G1, the same table may be near the chest, as shown in Figure~\ref{fig:embodiment_gap}. A retargeted motion that only matches human joint or keypoint motion will miss the object. The same issue appears when a robot crouches to pick up an object under a table while using the other hand for support: success depends on preserving the support contact, avoiding the table, and keeping the motion dynamically trackable. Collision avoidance alone does not resolve this ambiguity: several collision-free poses may exist, but only a small subset preserves the demonstrated contact and approach speed.

% \begin{figure}[t]
%     \centering
%     \begin{minipage}[t]{0.38\linewidth}
%         \centering
%         \includegraphics[width=\linewidth]{figures/table_height_mismatch2.png}
%         \caption{Relative table-height mismatch across embodiments: the same table is near waist height for the human but near chest height for the shorter G1.}
%         \label{fig:embodiment_gap}
%     \end{minipage}
%     \hfill
%     \begin{minipage}[t]{0.58\linewidth}
%         \centering
%         \includegraphics[width=\linewidth]{figures/retarget_comp_rebuttal.pdf}
%         \caption{Tabletop retargeting: red indicates penetration or discontinuity; green indicates a collision-free, continuous path.}
%         \label{fig:tabletop_retargeting}
%     \end{minipage}
% \end{figure}

\begin{figure}[t]
    \centering
    \includegraphics[width=\linewidth]{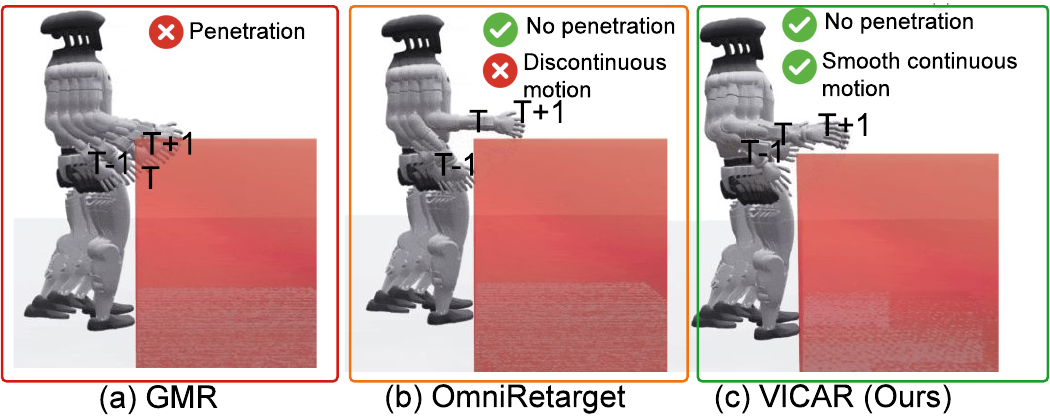}
    \caption{Tabletop retargeting: red indicates penetration or discontinuity; green indicates a collision-free, continuous path.}
    \label{fig:tabletop_retargeting}
\end{figure}
Many video-imitation pipelines rely on egocentric or robot-view demonstrations, but the web and everyday recordings contain a plethora of third-person human videos that can provide strong priors for humanoid whole-body motion learning. Recent systems show that monocular videos can provide scalable skill data for humanoids. VideoMimic~\cite{allshire2025videomimic} reconstructs human-scene geometry from casual videos and uses the resulting meshes and retargeted motions for contextual control. HDMI~\cite{weng2025hdmi} learns humanoid-object interaction policies from monocular RGB videos by co-tracking robot and object states. OmniRetarget~\cite{yang2025omniretarget} improves interaction-preserving retargeting by preserving human-object-terrain relations through an interaction mesh and hard kinematic constraints. These works make major progress, but our target setting places a different emphasis: we want the robot to preserve \emph{specific global contact poses} that may move at runtime, such as the ball-hit point in a table-tennis serve or the object position in a pickup task. The challenge is thus not only to reconstruct what the human did, but also to identify which interaction quantities must remain invariant after embodiment transfer.

We introduce \method{}, a video-to-humanoid learning pipeline centered on contact-aware retargeting. Given a third-person video, \method{} extracts human motion using GENMO/GEM~\cite{li2025genmo}, reconstructs task-relevant obstacle geometry using monocular depth and semantic segmentation. The selected regions are unprojected into a point cloud, voxelized into collision geometry, and scaled following the human-scene alignment ideas of VideoMimic~\cite{allshire2025videomimic}. We then retarget the motion in two stages: an initial OmniRetarget-style~\cite{yang2025omniretarget} pass gives a feasible robot reference, and a second optimization uses high-priority collision and global contact-pose penalties while preserving motion speed.

Our retargeted motions are not merely static references. They seed a motion generator conditioned on movable contact points that can be adjusted online during inference: for example, the ball-hit point in a table-tennis serve can move with the perceived ball position, and object-pick motions can be adapted to different object placements. A downstream policy then tracks the generated reference using proprioception and gravity observations, allowing deployment on hardware.

This paper makes the following contributions:
\begin{itemize}
    \item We formulate contact- and speed-preserving retargeting for third-person video imitation, using high-priority global contact-pose and collision penalties derived from monocularly reconstructed obstacles.
    \item We connect this retargeting stage to a contact-conditioned motion generator that supports dynamic contact locations and contact-point augmentation. 
    \item We evaluate the pipeline on ten diverse skills, including six table-tennis serves, object pickups above and below a table, bimanual box pickup and placement, and ladder climbing. All skills are deployed on hardware except ladder climbing.
    % Hardware trials achieve $21/30$, $10/10$, $7/10$, and $6/10$ successes on the four deployed task groups, respectively; ladder climbing remains simulation-only.
\end{itemize}

%% file: sections/02_related_work.tex
\section{Related Work}

\paragraph{Humanoid learning from human videos}
Human video is an attractive supervision source because it is easier to collect than teleoperation, motion capture, or robot demonstrations. Recent systems learn pose or locomotion from large video collections~\cite{luo2023perpetual,fu2024humanplus,he2025asap,mao2024massivevideos,allshire2025videomimic,yang2026zerowbc}, adapt robot-free videos to manipulation~\cite{li2024okami,lepert2025masquerade,shah2025mimicdroid,shi2026egohumanoid}, or target humanoid-object interaction~\cite{weng2025hdmi,zhao2025resmimic,wang2026humanx,wu2026sugar}. HDMI~\cite{weng2025hdmi} extracts human and object trajectories for co-tracking policies, while VideoMimic~\cite{allshire2025videomimic} reconstructs human-scene geometry for contextual skills. HumanX and SUGAR scale video-derived interaction learning~\cite{wang2026humanx,wu2026sugar}, but do not explicitly preserve task-specified global contact poses that can change at runtime. These systems primarily expand reference acquisition and policy learning; our focus is the intermediate geometric adaptation needed to keep sparse, task-defining contacts valid after embodiment transfer.

\paragraph{Human-to-humanoid retargeting}
Kinematic imitation uses retargeted human motion as reference data for characters and robots~\cite{peng2018deepmimic,wang2023physhoi,tessler2024maskedmimic,xu2026intermimic}. PHC~\cite{luo2023perpetual} fits SMPL-style pose to a robot embodiment, and GMR~\cite{araujo2025retargeting} matches human pose through relative keypoint geometry. Video-to-humanoid systems such as HumanPlus and ResMimic use retargeted references for tracking~\cite{fu2024humanplus,zhao2025resmimic}, but do not directly preserve global contacts or avoid task obstacles. OmniRetarget~\cite{yang2025omniretarget} preserves local human--object--terrain relations with an interaction mesh and hard joint, velocity, collision, and stance constraints. \method{} instead represents selected contacts as world-frame pose objectives and optimizes the whole clip to preserve source speed through contact. The distinction matters when collision avoidance admits several geometrically feasible solutions: in tabletop pickup, for example, entering and exiting the platform through different faces produces a discontinuous reference despite satisfying local clearance.

\paragraph{Scene reconstruction for robot learning}
Monocular depth and segmentation provide useful but noisy scene geometry~\cite{piccinelli2024unidepth,kirillov2023segmentanything,li2023semanticsam}. Video-to-robot pipelines also use body recovery, object reconstruction, 6D pose tracking, and registration to extract interaction priors~\cite{yang2026sam3dbody,gao2025sambody4d,sam3dteam2025sam3d,foundationposewen2024,besl1992method,wu2026sugar}. Similar to VideoMimic~\cite{allshire2025videomimic}, we unproject depth and resolve scale from the human-scene geometry, but reconstruct only semantic regions likely to constrain the robot. This task-focused reconstruction avoids requiring a complete scene model while retaining the surfaces needed for collision and contact reasoning.

\paragraph{Contact-conditioned motion generation}
Interactive humanoid policies often track reference robot and object states~\cite{weng2025hdmi,zhao2025resmimic} or reconstructed scene-conditioned references~\cite{allshire2025videomimic}. SUGAR~\cite{wu2026sugar} improves generalization with a hierarchical command generator after physics refinement. SceneBot~\cite{chen2026scenebot} and ContactMimic~\cite{li2026contactmimic} show that per-link or binary contact commands can condition the tracking policy directly when scene interaction is represented implicitly. This policy-level conditioning is sufficient when the controller can resolve the desired interaction from a motion and contact mode. In contrast, retargeting-layer anchors are needed when the world exogenously determines a specific $\SE(3)$ pose---such as a measured ball location or fixed ladder rung---that must first be embedded in the reference motion and interpolated across contact variants. \method{} places a contact-conditioned generator between retargeting and policy learning, separating geometric adaptation from low-level control.

\begin{figure*}[t]
    \centering
    \includegraphics[width=\textwidth]{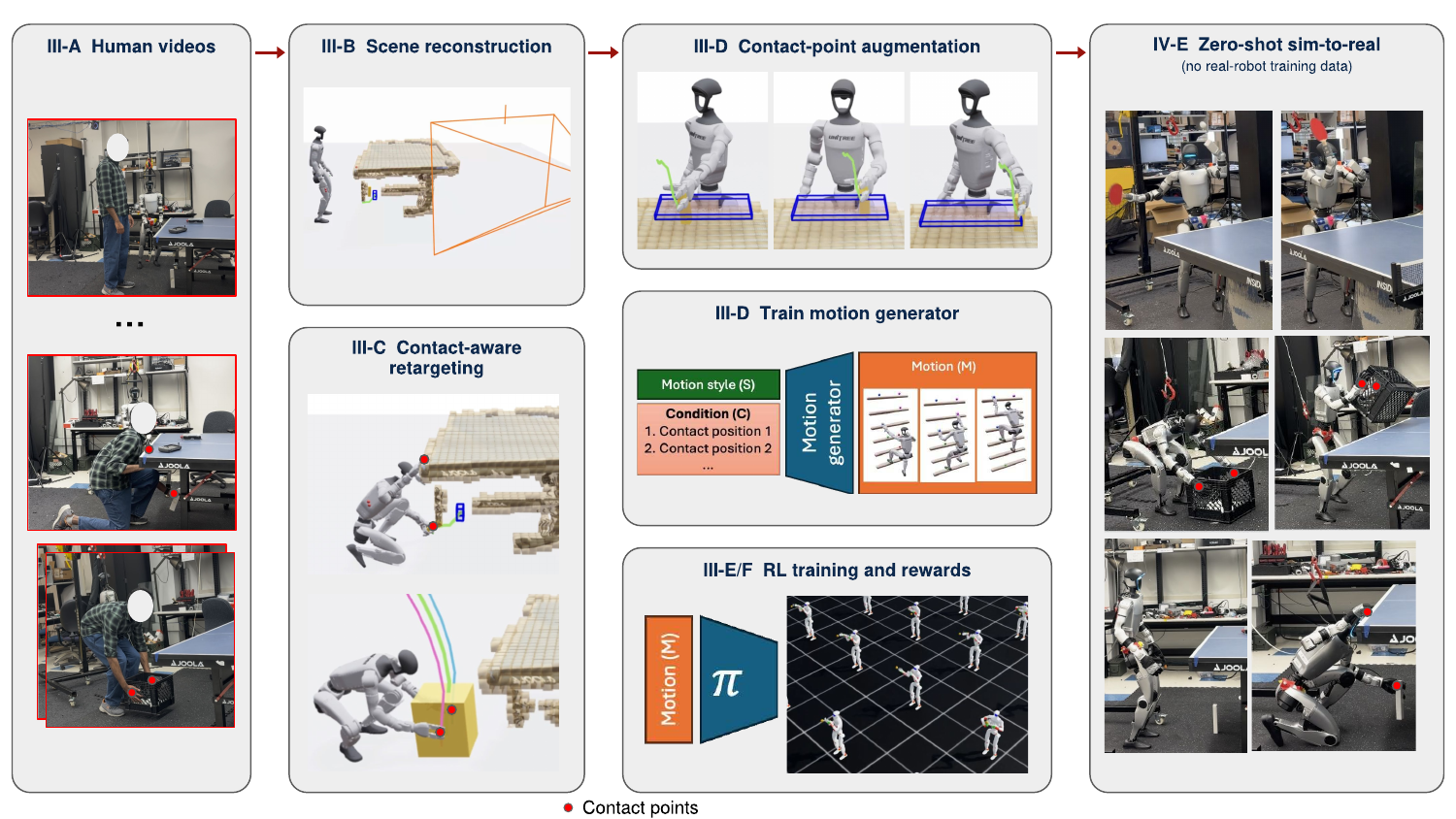}
    \caption{Overview of \method{}, with corresponding manuscript subsections labeled in the diagram. Deployment is zero-shot sim-to-real in the sense that training uses no real-robot data; motion capture supplies state observations at test time.}
    \label{fig:placeholder}
\end{figure*}

\begin{table}[t]
\centering
\small
\caption{Comparison of retargeting methods. ``Global contact poses'' denotes task-specified world-frame pose objectives}
\label{tab:retargeting_comparison}
\resizebox{\linewidth}{!}{
\begin{tabular}{@{}lcccccc@{}}
\toprule
Method &
\shortstack{Kin./collision\\handling} &
\shortstack{Object\\interaction} &
\shortstack{Terrain\\interaction} &
\shortstack{Data\\aug.} &
\shortstack{Preserve global\\contact poses} &
\shortstack{Preserve global\\relative speed} \\
\midrule
PHC~\cite{luo2023perpetual} & -- & \(\times\) & \(\times\) & \(\times\) & \(\times\) & \(\times\) \\
GMR~\cite{araujo2025retargeting} & -- & \(\times\) & \(\times\) & \(\times\) & \(\times\) & \(\times\) \\
VideoMimic~\cite{allshire2025videomimic} & Penalty & \(\times\) & \(\checkmark\) & \(\times\) & \(\times\) & \(\times\) \\
OmniRetarget~\cite{yang2025omniretarget} & Hard & \(\checkmark\) & \(\checkmark\) & \(\checkmark\) & Relative & \(\times\) \\
\method{} (Ours) & Penalty & \(\checkmark\) & \(\checkmark\) & \(\checkmark\) & \(\checkmark\) & \(\checkmark\) \\
\bottomrule
\end{tabular}
}
\end{table}

%% file: sections/03_method.tex
\section{Method}
\label{sec:method}

The input to \method{} is a monocular third-person RGB video of a human performing an interaction skill. The output is a policy that tracks a generated humanoid reference while allowing selected contact poses to vary at inference time. The pipeline has five stages.

\subsection{Human Motion and Contact Annotation}

We extract global human motion from the input video using GENMO/GEM~\cite{li2025genmo}. The resulting sequence provides source body poses, global root motion, and task-relevant body keypoints. For the current prototype, contact frames, contacted body parts, and contact-frame orientations are labeled manually. These annotations define tuples
\begin{equation}
    \mathcal{A} = \{(\tau_k, b_k, \hat{\mathbf{p}}_k, \hat{\mathbf{R}}_k)\}_{k=1}^{K},
\end{equation}
where $\tau_k$ is a frame, $b_k$ is a robot link or end-effector, and $(\hat{\mathbf{p}}_k,\hat{\mathbf{R}}_k)\in\R^3\times\SO(3)$ is the desired global contact pose. Thus, a \emph{contact} means that a specified robot body reaches a world-frame $\SE(3)$ pose at a specified timestep; it is a pose objective only and does not model force, contact mode, or sliding. In table tennis, for example, we annotate one pose for the ball toss and one for the racket--ball hit, and the hit pose can move online with the estimated ball state.

\subsection{Obstacle Reconstruction from Video}

The obstacle reconstruction stage identifies the portions of the scene that constrain the robot motion. We first estimate per-frame monocular depth using an off-the-shelf metric depth model: UniDepth~\cite{piccinelli2024unidepth}, and segment the image into semantic regions using Semantic-SAM ~\cite{kirillov2023segmentanything,li2023semanticsam}. We then query \gmini{}~\cite{openai2024gpt4omini} with the segmented frame and task description to select regions likely to be in contact with the human, such as a tabletop, table edge, floor patch, or support surface. 

For each selected region, we unproject depth pixels into camera-frame points and transform them into the task world frame. Following VideoMimic's use of human-scene alignment and metric scale~\cite{allshire2025videomimic}, we estimate a scale factor from the recovered human height from GENMO/GEM~\cite{li2025genmo} and estimated height from human semantic region depths. The selected points are filtered and converted into voxelized mesh obstacles:
\begin{equation}
    \mathbf{x}_{u,v} = \alpha \, \mathbf{T}_{w c} \, \mathbf{K}^{-1}
    \begin{bmatrix} u & v & 1 \end{bmatrix}^{\top} D(u,v),
\end{equation}
where $\mathbf{x}_{u,v}\in\R^3$, $D(u,v)\in\R_+$ is depth, $\mathbf{K}\in\GL(3)$ is the camera intrinsic matrix, $\mathbf{T}_{wc}\in\SE(3)$ is the camera-to-world transform, and $\alpha\in\R_+$ is the scale factor. In this stage, a voxel grid is regarded as a set of cuboid obstacles. We use voxel grid size of $0.05m$ side-length cuboids with a threshold of $5$ points to consider as occupied with obstacle.

\subsection{Contact- and Speed-Preserving Retargeting}

We retarget in two stages. First, we obtain an initial robot trajectory $\bar{q}_{1:T}$ using an OmniRetarget-style keypoint formulation~\cite{yang2025omniretarget} with obstacle penalties disabled. This gives a reasonable motion prior under robot joint limits. Second, we optimize the trajectory with high-priority contact and collision penalties while preserving the source motion's speed profile at the end-effectors.

\subsubsection{Stage 1: Interaction-Mesh Initialization}

The first pass follows OmniRetarget's interaction-mesh formulation but omits its obstacle-avoidance and penetration penalties. For keypoint $i$ at time $t$, the interaction mesh encodes local geometry using the Laplacian coordinate
\begin{equation}
L(\mathbf{p}_{t,i})=\mathbf{p}_{t,i}
-\sum_{j\in\mathcal{N}(i)}w_{ij}\mathbf{p}_{t,j},
\qquad w_{ij}=\frac{1}{|\mathcal{N}(i)|},
\label{eq:stage1_laplacian}
\end{equation}
where $\mathcal{N}(i)$ is the neighborhood of keypoint $i$. The deformation loss matches the source and robot Laplacian coordinates:
\begin{equation}
E_L(q_t)=\sum_i\left\|
L(\mathbf{p}^{\mathrm{src}}_{t,i})-
L(\mathbf{p}^{\mathrm{rob}}_{t,i}(q_t))
\right\|_2^2.
\label{eq:stage1_deformation}
\end{equation}
We solve this motion-fitting problem over the active joints with temporal regularization and joint feasibility:
\begin{align}
\bar{q}_t=\arg\min_{q_t}\quad
&E_L(q_t)+\|q_t-\bar{q}_{t-1}\|_Q^2
\label{eq:stage1_optimization}\\
\text{subject to}\quad
&q_{\min}\leq q_t\leq q_{\max}, \nonumber\\
&v_{\min}\Delta t\leq q_t-q_{t-1}\leq v_{\max}\Delta t.
\nonumber
\end{align}
Unlike the full OmniRetarget program, Eq.~\eqref{eq:stage1_optimization} does not include a signed-distance penalty. In implementation, this is equivalent to running the batched joint-angle solver with an empty obstacle list. The resulting $\bar{q}_{1:T}$ supplies a motion prior without prematurely forcing the robot around potentially mismatched scene geometry.

\subsubsection{Stage 2: Contact- and Speed-Preserving Refinement}

Let $q_t\in\mathcal{Q}$ denote the robot floating-base and joint configuration at time $t$. For a set of key links $\mathcal{B}$, forward kinematics gives world positions $\mathbf{p}^{b,R}_t(q_t)\in\R^3$ and orientations $\mathbf{R}^{b,R}_t(q_t)\in\SO(3)$. The source human positions and orientations are denoted $\mathbf{p}^{b,H}_t\in\R^3$ and $\mathbf{R}^{b,H}_t\in\SO(3)$ after semantic correspondence between human and robot keypoints. We solve
\begin{align}
q^\star_{1:T} = \arg\min_{q_{1:T}}\quad
&\mathcal{L}_p(q)+\lambda_R\mathcal{L}_R(q)
+\lambda_s\mathcal{L}_s(q) \nonumber\\
&{}+\lambda_{\mathrm{col}}\mathcal{L}_{\mathrm{col}}(q)
+\lambda_a\mathcal{L}_{\mathrm{anchor}}(q)
\label{eq:retarget_objective}
\end{align}
\begin{align}
\mathcal{L}_p(q)
= {}& \sum_{b \in \mathcal{B}}\sum_{t=2}^{T}
\Big(\big\|\Delta \mathbf{p}^{b,H}_t\big\|_2 \nonumber\\
&\hspace{4.5em}{}-
\big\|\Delta \mathbf{p}^{b,R}_t(q)\big\|_2\Big)^2, \nonumber\\
\mathcal{L}_R(q)
= {}& \sum_{b \in \mathcal{B}}\sum_{t=2}^{T}
\Big(d_R\!\left(\Delta \mathbf{R}^{b,H}_t,\mathbf{I}_3\right) \nonumber\\
&\hspace{3.2em}{}-
d_R\!\left(\Delta \mathbf{R}^{b,R}_t(q),\mathbf{I}_3\right)\Big)^2, \nonumber\\
\mathcal{L}_s(q)
= {}& \sum_{t=2}^{T}\|q_t-q_{t-1}\|_2^2, \nonumber\\
\mathcal{L}_{\mathrm{col}}(q)
= {}& \sum_{t,j}\left[\max\!\left(0,-\phi_j(q_t)\right)\right]^2, \nonumber\\
\mathcal{L}_{\mathrm{anchor}}(q)
= {}& \sum_{k=1}^{K}\left(\|\mathbf{e}^p_k\|_2^2
+\beta_R(e^R_k)^2\right), \label{eq:anchors}\\[-0.25em]
\text{subject to}\quad
& q_{\min} \leq q_t \leq q_{\max},\quad\forall t, \label{eq:jointlimits}\\
& v_{\min}\Delta t \leq q_t-q_{t-1} \leq v_{\max}\Delta t,
\quad\forall t. \label{eq:vellimits}
\end{align}
Here $\Delta \mathbf{p}^{b,\cdot}_t=\mathbf{p}^{b,\cdot}_t-\mathbf{p}^{b,\cdot}_{t-1}$, $\Delta \mathbf{R}^{b,\cdot}_t=\mathbf{R}^{b,\cdot}_t(\mathbf{R}^{b,\cdot}_{t-1})^\top$, and $d_R$ is the geodesic distance on $\SO(3)$. The anchor errors are $\mathbf{e}^p_k=\mathbf{p}^{b_k,R}_{\tau_k}(q_{\tau_k})-\hat{\mathbf{p}}_k$ and $e^R_k=d_R(\mathbf{R}^{b_k,R}_{\tau_k}(q_{\tau_k}),\hat{\mathbf{R}}_k)$. The first two terms preserve translational and angular speed, while $\mathcal{L}_s$ smooths the trajectory. The collision and anchor terms are high-priority Lagrangian (KKT-style) penalties: their large weights prioritize clearance and the desired global contact pose, but gradient descent does not guarantee exact equality. If an annotation conflicts with feasible geometry, the optimizer returns the nearest feasible pose and we report its residual explicitly.

In practice, we optimize the full trajectory over each task's active joints. Collision penalties are computed against cuboid or mesh obstacles, anchor costs act over a small temporal window, and joint clamps enforce limits. 

\subsection{Contact-Conditioned Motion Generator}
\label{sec:contact_generator}

The optimized trajectory dataset is used to train a motion generator $G_\theta$. At each step, the generator receives a motion style vector $\mathbf{s}$, movable contact conditions $\mathbf{c}$, and a temporal context, and predicts reference robot and object motion:
\begin{equation}
    (\hat{\mathbf{m}}_{t:t+\Delta}, \hat{\mathbf{o}}_{t:t+\Delta})
    =
    G_\theta(\mathbf{s}, \mathbf{c}, \mathbf{m}_{\leq t}, \mathbf{o}_{\leq t}).
\end{equation}
The style vector is a one-hot identifier for the source video or motion family. For the six table-tennis serves, this vector selects among simple forehand, simple backhand, forehand chop, backhand chop, forehand side spin, and backhand side spin. The contact vector stores task-dependent contact positions: toss and hit points for serving, object pickup contact pose for manipulation, and hand/foot support contacts for ladder climbing.

\paragraph{Contact-point augmentation.}
For serving, we augment the racket--ball hit point on a 3D grid around the demonstrated location while preserving the toss and serve style; one-hot style vectors and per-style contact ranges represent the six serves. For object pickup, we vary the object pose within task-specific ranges. Bimanual variants keep both hands synchronized with the box, while ladder variants preserve the sequential hand and foot support anchors. The resulting optimized variants train the generator and policy over movable contact poses.
% \todo{Add network architecture, loss terms, dataset size, and training hyperparameters.}

\subsection{RL Environment and Training}
\label{sec:rl_training}

We train the tracking policy in Isaac Lab with a Unitree G1 model and joint-position actions. The simulator uses a $0.005$ s physics step and action decimation of $4$, yielding a nominal $50$ Hz policy rate; the low-frequency variant runs at $25$ Hz. Each episode follows the motion length. We use $4096$ parallel environments with $2.5$ m spacing. 
% Most tasks use a ground plane together with the obstacles extracted from video. 
% For ladder climbing, where monocular reconstruction is unreliable near thin step edges, we instead use six cuboid steps with $0.30$ m riser height, $0.076$ m tread, and $1.0$ m depth.

\paragraph{Policy observations and actions.}
The observation concatenates the reference command; motion-anchor position and orientation in the robot anchor frame; body positions and orientations relative to the anchor; base linear and angular velocity; relative joint position and velocity; and the previous action. Observation corruption uses uniform noise with ranges $\pm0.25$ m for anchor position, $\pm0.05$ for anchor orientation, $\pm0.5$ m/s for base linear velocity, $\pm0.2$ rad/s for base angular velocity, $\pm0.01$ rad for joint position, and $\pm0.5$ rad/s for joint velocity. Actions are joint-position targets about the default pose, scaled per actuator using effort limits and stiffness.

\paragraph{PPO and actor--critic architecture.}
We use PPO~\cite{schulman2017ppo} with empirical observation normalization. The actor and critic are MLPs with hidden dimensions $[512,256,128]$, ELU activations, and initial action-noise standard deviation $1.0$. PPO uses $24$ steps per environment, $5$ learning epochs, $4$ minibatches, an adaptively scheduled learning rate of $10^{-3}$, clipping parameter $0.2$, entropy coefficient $0.005$, value-loss coefficient $1.0$, discount $\gamma=0.99$, GAE $\lambda=0.95$, desired KL $0.01$, and maximum gradient norm $1.0$. Training runs for up to $30{,}000$ iterations, with checkpoints every $500$ iterations.

\paragraph{Domain randomization.}
Following the minimal randomization strategy of OmniRetarget~\cite{yang2025omniretarget}, we randomize robot and contact parameters without task-specific curricula. We sample static friction in $[0.3,1.6]$, dynamic friction in $[0.3,1.2]$, restitution in $[0,0.5]$, joint default offsets in $\pm0.01$ rad, and torso center-of-mass offsets of $\pm0.025$ m in $x$ and $\pm0.05$ m in $y,z$. Random pushes occur every $1$--$3$ s, with linear velocity ranges $x,y\in[-0.5,0.5]$ m/s and $z\in[-0.2,0.2]$ m/s, and angular velocity ranges of $[-0.52,0.52]$ rad/s in roll/pitch and $[-0.78,0.78]$ rad/s in yaw.

% At deployment, the motion generator updates the contact conditions online and the policy tracks the resulting reference on the Unitree G1.

\subsection{Rewards}
\label{sec:rewards}

Following OmniRetarget~\cite{yang2025omniretarget} and BeyondMimic~\cite{zeng2025beyondmimic}, the reward combines anchor, body-pose, body-velocity, and optional object tracking with action-rate, soft joint-limit, and undesired-contact penalties:
\begin{equation}
    r_t =
    \sum_i w_i r_i
    - w_a\|\mathbf{a}_t-\mathbf{a}_{t-1}\|_2^2
    - w_j r_{\mathrm{joint}}
    - w_c r_{\mathrm{contact}} .
    \label{eq:policy_reward}
\end{equation}
Tracking terms use
\begin{equation}
    r_{\mathrm{exp}}(\mathbf{e};\sigma)
    =
    \exp\!\left(-\frac{\|\mathbf{e}\|_2^2}{\sigma^2}\right),
    \label{eq:tracking_reward}
\end{equation}
where orientation errors use the geodesic quaternion distance. Table~\ref{tab:policy_rewards} gives the weights used for the G1 policy.

\begin{table*}[t]
    \centering
    \scriptsize
    \setlength{\tabcolsep}{3pt}
    \caption{G1 tracking and regularization rewards; tracking terms use Eq.~\eqref{eq:tracking_reward}.}
    \begin{tabular}{p{0.20\textwidth}p{0.07\textwidth}p{0.06\textwidth}p{0.58\textwidth}}
        \toprule
        Term & Weight & Std. & Description \\
        \midrule
        Global anchor position & $0.5$ & $0.3$ & Tracks torso/anchor world position. \\
        Global anchor orientation & $0.5$ & $0.4$ & Tracks anchor orientation. \\
        Relative body position & $1.0$ & $0.3$ & Tracks selected body-link positions relative to the anchor. \\
        Relative body orientation & $1.0$ & $0.4$ & Tracks selected body-link orientations. \\
        Body linear velocity & $1.0$ & $1.0$ & Tracks selected body-link linear velocities. \\
        Body angular velocity & $1.0$ & $\pi$ & Tracks selected body-link angular velocities. \\
        Action rate & $-0.1$ & -- & Penalizes $\|\mathbf{a}_t-\mathbf{a}_{t-1}\|_2^2$. \\
        Soft joint limit & $-10.0$ & -- & Penalizes excursions beyond soft joint-position limits. \\
        Undesired contact & $-0.1$ & -- & Penalizes non-allowed links with contact force above $1$ N; feet and wrists are excluded. \\
        \bottomrule
    \end{tabular}
    \label{tab:policy_rewards}
\end{table*}
% \vspace{-5mm}

Episodes terminate on timeout or when anchor-height error exceeds $0.25$ m, projected-gravity error exceeds $0.8$, or ankle/wrist height error exceeds $0.25$ m.

%% file: sections/04_experiments.tex
\section{Experiments}
\label{sec:experiments}

Our evaluation is designed to answer four questions: (i) whether contact-aware retargeting reduces contact error and collision while preserving motion smoothness; (ii) whether it outperforms OmniRetarget-style retargeting on tasks where global contact pose matters; (iii) whether contact-conditioned generation can adapt the same motion to novel contact points; and (iv) whether the generated references successfully transfer to the real Unitree G1 robot.

\subsection{Tasks}

We evaluate our approach on ten third-person video skills, summarized in Table~\ref{tab:tasks}. A \emph{skill} is one source video and its contact-conditioned motion family; the six serves are distinct skills with different styles. The first six skills have movable toss and ball-hit points, the next three vary object-pick poses, and ladder climbing has sequential hand and foot supports.

% \vspace{-8mm}
\begin{table*}[t]
    \centering
    \small
    \begingroup
    \setlength{\tabcolsep}{3pt}
    \label{tab:tasks}
    \caption{Evaluation task suite. } %Replace contact ranges and trial counts with final values.}
    \begin{tabular}{@{}p{0.04\linewidth}p{0.28\linewidth}p{0.33\linewidth}p{0.27\linewidth}@{}}
    \toprule
    \# & Skill & Preserved Contacts & Movable Contacts \\
    \midrule
    1 & Simple forehand serve & Toss & Racket hit pose \\
    2 & Simple backhand serve & Toss & Racket hit pose \\
    3 & Forehand chop serve & Toss & Racket hit pose \\
    4 & Backhand chop serve & Toss & Racket hit pose \\
    5 & Forehand side-spin serve & Toss & Racket hit pose \\
    6 & Backhand side-spin serve & Toss & Racket hit pose \\
    7 & Pick object from tabletop & - & Object grasp pose \\
    8 & Pick object under table  & Right-hand table-edge support & Left-hand object grasp pose \\
    9 & Bimanual box pickup from ground and placement on table & Synchronized two-hand box grasp & - \\
    10 & Ladder climbing & Hand and foot rung supports & - \\
    \bottomrule
    \end{tabular}
    \endgroup
\end{table*}
% \vspace{-5mm}

\subsection{Baselines and Metrics}

We compare \method{} against OmniRetarget~\cite{yang2025omniretarget}, GMR~\cite{araujo2025retargeting}, PHC~\cite{luo2023perpetual}, VideoMimic~\cite{allshire2025videomimic}, and an ablation without speed preservation. Identical contact-pose penalty terms are added post-hoc to every baseline, and every method uses the same anchor-pose augmentations. Downstream tracking policies use identical PPO hyperparameters, training steps, domain randomization, and rewards. The comparison therefore isolates retargeting quality rather than contact formulation, augmentation, or policy compute.
We report the following metrics:
\begin{itemize}
    % \item \textbf{Contact position error}: distance between robot contact link and desired global contact point at annotated contact frames.
    % \item \textbf{Contact orientation error}: geodesic error between contact link orientation and the desired contact frame.
    \item \textbf{Penetration and residuals}: mean signed penetration depth; mean position residual at annotated anchors; and infeasibility rate, defined as the fraction of frames with penetration greater than $1$ cm.
    \item \textbf{Smoothness}: mean joint jerk and maximum joint velocity in motion (to measure discontinuity in motions). For optimized joint trajectories $q^{(n)}_{1:T}$ over $N$ samples and $J$ joints, mean joint jerk is computed as the average absolute third finite difference,
    \begin{equation}
        \begin{aligned}
        J_{\mathrm{jerk}}
        &= \frac{1}{N(T-3)J}
        \sum_{n=1}^{N}\sum_{t=1}^{T-3}\sum_{j=1}^{J}
        \left|q^{(n)}_{t+3,j}-3q^{(n)}_{t+2,j}\right. \\
        &\hspace{5.8em}\left.{}+3q^{(n)}_{t+1,j}-q^{(n)}_{t,j}\right|,
        \end{aligned}
        \label{eq:smoothness}
    \end{equation}
    reported in rad/step$^3$.
    \item \textbf{Speed preservation}: error between source and retargeted keypoint displacement magnitudes of end-effectors (wrist and feet). We compute
    \begin{equation}
        \begin{aligned}
        E_{\mathrm{speed}}
        &= \frac{1}{|\mathcal{B}_{\mathrm{ee}}|(T-1)}
        \sum_{b \in \mathcal{B}_{\mathrm{ee}}}\sum_{t=2}^{T} \\
        &\quad \left(
        \left\|\Delta \mathbf{p}^{b,H}_t\right\|_2
        -\left\|\Delta \mathbf{p}^{b,R}_t\right\|_2
        \right)^2,
        \end{aligned}
        \label{eq:speed_preservation_metric}
    \end{equation}
    where $\mathcal{B}_{\mathrm{ee}}$ is the set of evaluated end-effectors and $\Delta \mathbf{p}^{b,\cdot}_t=\mathbf{p}^{b,\cdot}_t-\mathbf{p}^{b,\cdot}_{t-1}$.
    \item \textbf{Downstream RL Policy success rate}: success rate of trained policy in simulation over 1000 trials.
\end{itemize}

\subsection{Retargeting Results}

% \placeholderfigure{1.35in}{Qualitative comparison with other retargeting techniques, similar to omniretarget}

% \begin{figure}[htbp]
%     \centering
%     \begin{minipage}[t]{0.45\linewidth}
%         \centering
%         \includegraphics[width=\linewidth]{figures/aug.pdf}
%         \caption{Motion augmentation by contact points.}
%         \label{fig:aug}
%     \end{minipage}
%     \hfill
%     \begin{minipage}[t]{0.48\linewidth}
%         \centering
%         \includegraphics[width=\linewidth]{figures/success_rate_icra.pdf}
%         \caption{Success rate at held-out in-distribution contact points versus the number of augmented trajectories.}
%         \label{fig:success_rate}
%     \end{minipage}
% \end{figure}

\begin{figure}[t]
    \centering
    \includegraphics[width=0.72\linewidth]{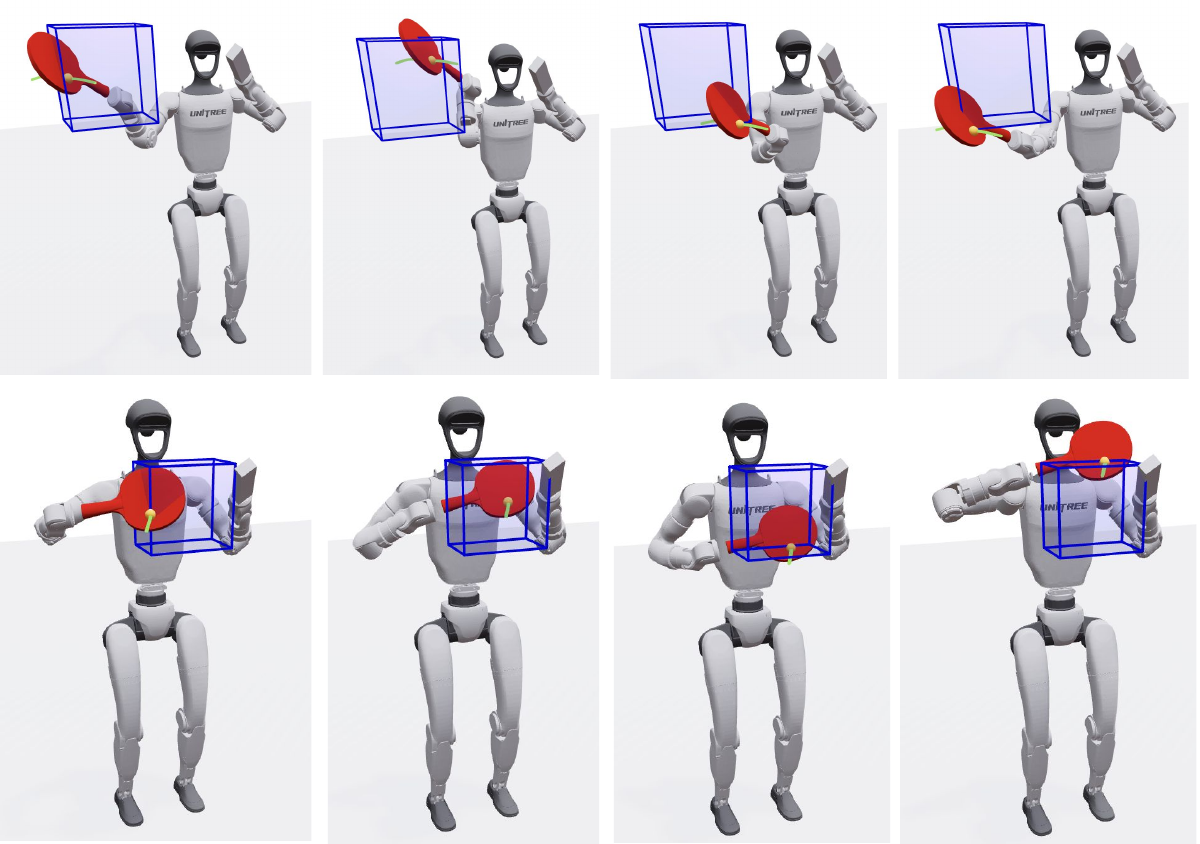}
    \caption{Motion augmentation by contact points.}
    \label{fig:aug}
\end{figure}

\begin{figure}[t]
    \centering
    \includegraphics[width=0.78\linewidth]{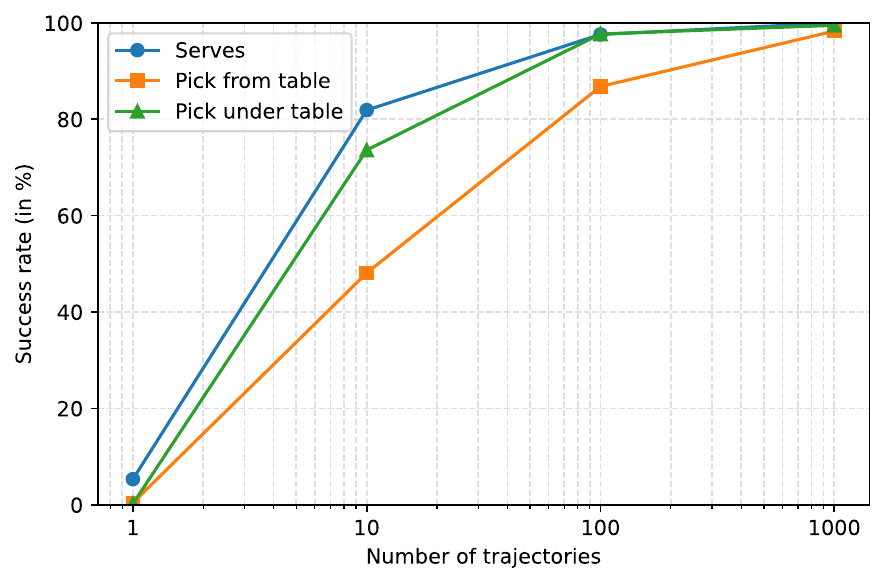}
    \caption{Success rate at held-out in-distribution contact points versus the number of augmented trajectories.}
    \label{fig:success_rate}
\end{figure}
% \placeholderfigure{1.35in}{Qualitative comparison on the tabletop object-pick task. The expected figure should show OmniRetarget entering or exiting the platform through different faces, causing a discontinuous correction, while \method{} preserves the object contact and clears the platform smoothly.}{fig:tabletop_failure}{Placeholder for task 7 qualitative comparison: OmniRetarget vs. \method{} on table pickup.}

Table~\ref{tab:retargeting} reports the quantitative comparison. Table-tennis success depends strongly on augmentation and runtime contact estimation, but a smooth reference remains a prerequisite: GMR attains $97.2\%$ versus \method{}'s $99.8\%$. The clearest gains occur on obstacle-critical tasks, where downstream success rises from $76.2\%$ to $99.8\%$ for tabletop pickup and from $56.0\%$ to $95.5\%$ for ladder climbing relative to OmniRetarget. In tabletop pickup, OmniRetarget's initial reference enters and exits the platform through different faces, so a later non-penetration correction becomes discontinuous; \method{} instead preserves contact and speed while finding a smooth path (Figure~\ref{fig:tabletop_retargeting}). 
% Removing speed preservation raises speed error to approximately $0.35$ m/step and reduces success to $63.2\%$ and $47.3\%$ on these two tasks. 
% The abrupt foothold change visible during ladder execution arises in the downstream tracking policy, not in the retargeted trajectory, whose jerk remains low.

\begin{table*}[t]
    \centering
    \small
    \caption{Retargeting and downstream tracking results. Cells follow the group-header order. Anchor residual and infeasibility (frames with penetration $>1$ cm) are reported for \method{}; -- denotes unreported measurements. The speed ablation is evaluated on the two obstacle-critical tasks.}
    \label{tab:retargeting}
    \resizebox{\textwidth}{!}{
    \begin{tabular}{@{}lccccccc@{}}
    \toprule
    \multicolumn{8}{c}{\textit{Table-tennis serves / Bimanual pick-place / Pick under table}} \\
    \cmidrule(lr){1-8}
    Method & \shortstack{Pen. (cm)\\$\downarrow$} & \shortstack{Anchor res. (cm)\\$\downarrow$} & \shortstack{Infeas. (\%)\\$\downarrow$} & \shortstack{Avg jerk (rad/step$^3$)\\$\downarrow$} & \shortstack{Max vel. (rad/step)\\$\downarrow$} & \shortstack{Speed err. (m/step)\\$\downarrow$} & \shortstack{RL success (\%)\\$\uparrow$} \\
    \midrule
    PHC~\cite{luo2023perpetual} & \textbf{0.00}/3.54/0.75 & -- & -- & 0.059/\textbf{.0020}/\textbf{.0016} & 1.496/\textbf{.205}/\textbf{.096} & .014/.073/.013 & 41.0/0.0/0.0 \\
    GMR~\cite{araujo2025retargeting} & \textbf{0.00}/4.21/7.43 & -- & -- & .033/.0049/.0045 & .831/1.159/.523 & .0018/.397/.023 & 97.2/52.4/46.8 \\
    VideoMimic~\cite{allshire2025videomimic} & \textbf{0.00}/.53/.37 & -- & -- & .033/.0051/.0054 & .894/1.204/.614 & .0018/.464/.120 & 96.3/99.7/94.6 \\
    OmniRetarget~\cite{yang2025omniretarget} & \textbf{0.00}/\textbf{.03}/.06 & -- & -- & .062/.0046/.0020 & 1.597/.833/.176 & .014/.946/.013 & \textbf{99.8}/\textbf{100}/98.1 \\
    \method{} & \textbf{0.00}/.04/\textbf{.01} & .12/.42/.52 & 0.0/0.0/0.0 & \textbf{.032}/.0026/.0024 & \textbf{.787}/.249/.098 & \textbf{.000}/\textbf{.0014}/\textbf{.00019} & \textbf{99.8}/\textbf{100}/\textbf{99.2} \\
    \midrule
    \multicolumn{8}{c}{\textit{Pick from table / Ladder climbing}} \\
    \cmidrule(lr){1-8}
    Method & \shortstack{Pen. (cm)\\$\downarrow$} & \shortstack{Anchor res. (cm)\\$\downarrow$} & \shortstack{Infeas. (\%)\\$\downarrow$} & \shortstack{Avg jerk (rad/step$^3$)\\$\downarrow$} & \shortstack{Max vel. (rad/step)\\$\downarrow$} & \shortstack{Speed err. (m/step)\\$\downarrow$} & \shortstack{RL success (\%)\\$\uparrow$} \\
    \midrule
    PHC~\cite{luo2023perpetual} & 8.52/.54 & -- & -- & \textbf{.0012}/\textbf{.0043} & \textbf{.086}/\textbf{.195} & \textbf{.0015}/.278 & 0.0/0.0 \\
    GMR~\cite{araujo2025retargeting} & 8.15/1.32 & -- & -- & .0019/.0091 & .40/.748 & .013/.297 & 11.2/0.0 \\
    VideoMimic~\cite{allshire2025videomimic} & .21/.85 & -- & -- & .0037/.0096 & .42/.728 & .058/.310 & 62.2/0.0 \\
    OmniRetarget~\cite{yang2025omniretarget} & .02/.08 & -- & -- & .0032/.0078 & .37/.530 & .35/.419 & 76.2/56.0 \\
    \method{} & \textbf{.00}/\textbf{.05} & .19/1.5 & 0.0/2.6 & .0024/.0087 & .10/.303 & .027/\textbf{.012} & \textbf{99.8}/\textbf{95.5} \\
    % \method{} w/o speed & -- & -- & -- & -- & -- & $\sim0.35$ (both) & 63.2/47.3 \\
    \bottomrule
    \end{tabular}
    }
\vspace{-2mm}
\end{table*}

\vspace{-1mm}
\subsection{Contact-Point Generalization}

Figure~\ref{fig:success_rate} reports held-out contact-point performance as the number of augmented trajectories increases; the corresponding augmentation procedure is described in Section~\ref{sec:contact_generator}.

% \placeholderfigure{1.3in}{Contact augmentation for table-tennis serving. Replace with a grid of generated hit points, overlaid racket trajectories, and hardware stills for several contact points.}{fig:serve_aug}{Placeholder for serve augmentation: contact grid, generated racket trajectories, and hardware deployment snapshots.}

\vspace{-1mm}
\subsection{Hardware Deployment}

We deploy the policies on hardware. No real-robot training data are used, so deployment is zero-shot sim-to-real in that specific sense, not visual sim-to-real. The policy tracks generated references using proprioception and projected gravity, while an external motion-capture system supplies the global ball, object positions used to update contact conditions. The Dex3-1 hand is used for object pickup. Hardware success is $21/30$ ($70.0\%$) for table-tennis serves, $10/10$ ($100\%$) for tabletop pickup, $7/10$ ($70.0\%$) for under-table pickup, and $6/10$ ($60.0\%$) for bimanual box pickup. 
% On tabletop pickup, the matched OmniRetarget hardware baseline succeeds in $2/10$ trials versus \method{}'s $10/10$.

%% file: sections/05_limitations_conclusion.tex
% \section{Limitations}
% \label{sec:limitations}

% \method{} is a step toward learning interactive humanoid behavior from third-person videos, but the current system has several important limitations. First, the recovered motion must be retargetable to the Unitree G1. Motions requiring unreachable contacts, unsupported balance states, or dexterous manipulation beyond the robot hand model can still fail. Second, contact frames, contacted body parts, and desired contact orientations are manually labeled. The perception stack identifies likely contacted regions, but the final temporal contact schedule is not yet fully automated. Third, the deployed robot does not use vision in the current version: it must start from the same relative pose as the human demonstration or be given the contact pose externally. Fourth, obstacle reconstruction is only as good as the maonocular depth, segmentation, scale alignment, and semantic contact-region selection; thin structures and occluded supports remain challenging. Finally, the contact-conditioned motion generator currently relies on curated videos and manually defined augmentation ranges, so the system has not yet been evaluated as an open-ended internet-video learner.

\vspace{-1mm}
\section{Limitations}

This framework has several limitations. Contact frames, body parts, and $\SE(3)$ poses are manually annotated, requiring approximately $20$--$30$ minutes per skill; new task families still require per-task annotation and scene reconstruction. Motions with unreachable contacts, unsupported balance states, or dexterous manipulation beyond the hand model can fail. Ladder hardware deployment failed because the same blind tracking policy stumbled, even though the retargeted trajectory was smooth. Deployment also requires substantial motion-capture engineering to recover global pelvis, object, and ladder poses. Finally, robustness to monocular depth errors and Semantic-SAM or \gmini{} hallucinations is outside our present contribution. Future work will automate contact labeling, integrate onboard perception, and train vision-conditioned policies in reconstructed simulation twins.

\vspace{-1mm}
\section{Conclusion}

We introduced \method{}, a pipeline for learning humanoid interactive skills from third-person videos through contact-aware retargeting. The core idea is to make the useful part of a human demonstration explicit: not only the pose sequence, but also the global contacts that make the task work. By reconstructing task-relevant obstacles, preserving contact poses and speed during retargeting, and training a generator conditioned on movable contact points, \method{} can adapt video-derived skills to new object or ball positions before a tracking policy executes them on hardware. The current results span ten contact-rich motions, from table-tennis serves to pickup tasks and ladder climbing.